# Decision Making with Linear Constraints on Probabilities

Michael Pittarelli
SUNY College of Technology
Utica, NY 13504-3050

## 1. INTRODUCTION

Techniques for decision making with knowledge of linear constraints on condition probabilities are examined. These constraints arise naturally in many situations: upper and lower condition probabilities are known; an ordering among the probabilities is determined; marginal probabilities or bounds on such probabilities are known, e.g., data are available in the form of a probabilistic database (Cavallo and Pittarelli, 1987a); etc. Standard situations of decision making under risk and uncertainty may also be characterized by linear constraints. Each of these types of information may be represented by a convex polyhedron of numerically determinate condition probabilities. A uniform approach to decision making under risk, uncertainty, and partial uncertainty based on a generalized version of a criterion of Hurwicz is proposed. Methods for processing marginal probabilities to improve decision making using any of the criteria discussed are presented.

## 2. DECISIONS WITH MARGINAL PROBABILITIES

### 2.1 An Example

Suppose that it is known that a randomly selected object from a given container has a probability 0.7 of being black and 0.3 of being white, and a probability 0.6 of being spherical and 0.4 of being cylindrical. Suppose that nothing is known about the joint distribution of color and shape (e.g., independence of the two variables), but that one is forced to predict which of the four possible combinations of shape and color is possessed by a randomly selected object with utilities given by the decision matrix:

|          | $s_{BS}$ | $s_{BC}$ | $s_{WS}$ | $s_{WC}$ |
|----------|------|------|------|------|
| $a_{BS}$ | 10   | -5   | -5   | -5   |
| $a_{BC}$ | -1   | 10   | -1   | -1   |
| $a_{WS}$ | -50  | -50  | 540  | -50  |
| $a_{WC}$ | -1   | -1   | -1   | 10   |

(where $a_{\gamma\delta}$ and $s_{\gamma\delta}$ denote, respectively, the act of guessing that, and the state in which, the object has color $\gamma$ and shape $\delta$).

There are infinitely many joint distributions compatible with the marginal information, the convex polyhedron of solutions to the constrained system of linear equations:

$p(BS)+p(BC) = 0.7$

$p(WS)+p(WC) = 0.3$

$p(BS)+p(WS) = 0.6$

$p(BC)+p(WC) = 0.4$

$p(.) \geq 0.$

Such systems rarely have unique solutions. When they do, the uniquely determined joint distribution may be used to maximize expected utility in the standard Bayesian manner. Otherwise, a strict Bayesian is compelled to select, somehow, a single distribution from the infinite set of joint distributions compatible with the marginals.

### 2.2 (Maximum Entropy) Estimation

Appeal is frequently made in such cases to the *principle of maximum entropy* (PME). As stated by Jaynes: "when we make inferences based on incomplete information, we should draw them from that probability distribution that has the maximum entropy permitted by the information that we do have" (1982, p. 940). A PME decision rule may be formulated as follows. $S=\{s_1,\ldots,s_n\}$ is the set of relevant states of nature, $A=\{a_1,\ldots,a_m\}$ is the set of actions under consideration, and $U=\{u_{11},\ldots,u_{1n},u_{21},\ldots,u_{mn}\}$ is the set of utilities (expressed in "utiles"), where $u_{ij}$ is a function of $a_i$ and $s_j$. Let K denote the convex set of possible joint distributions. Then an action $a_k$ should be selected such that

$$\sum_{j=1}^{n} p^*(s_j)u_{kj} = \max_{i\in\{1,\dots,m\}} \sum_{j=1}^{n} p^*(s_j)u_{ij}$$

where $p^*$ is the (unique) maximum entropy element of K:

$$-\sum_{j=1}^{n} p^*(s_j)\log p^*(s_j) = \max_{p\in K} -\sum_{j=1}^{n} p(s_j)\log p(s_j).$$

For the example, the maximum entropy joint distribution is

| C | S | $p^*(\cdot)$ |
|---|---|---|
| B | S | 0.42 |
| B | C | 0.28 |
| W | S | 0.18 |
| W | C | 0.12 |

and the action maximizing expected utility is $a_{WS}$. Calculation of $p^*$ is discussed by Cavallo and Klir (1981) and by Cheeseman (1983).

A number of arguments have been advanced in support of PME as an estimation technique and, taken together, they are somewhat compelling. However, each is subject to more or less serious qualification (Frieden, 1985; Seidenfeld, 1986; Pittarelli, 1988). For example, Jaynes' *concentration theorem* (1979, 1982) has been interpreted as stating that the maximum entropy element of the set K is the distribution most likely to be the (single) "true" but unknown joint distribution compatible with the known marginals (Klir and Way, 1985). However, the result depends on probabilities being derived as relative frequencies; is an approximation (holds as $N\to\infty$, where N is the number of observations); and states merely that (in the limit) a fraction of the sequences of joint observations compatible with the marginals give rise to relative frequency distributions with entropy within $\Delta H$ of the entropy of $p^*$, where $2N\Delta H$ is distributed as $\chi^2$. The theorem really says nothing about the probability of occurrence of any of these relative frequency distributions. Outcomes associated with high-entropy distributions may be inherently less likely to arise, unnatural though this may seem to most people (accustomed to examples involving equally likely outcomes — rolls of fair die, tosses of fair coins, etc.).

## 3. ALTERNATIVES TO ESTIMATION

Several strategies for decision making from knowledge of marginal probabilities (and from other types of information determining a convex set of joint distributions — bounds on components $p(s_j)$, or an ordering among them) have been developed that do not require estimation of joint probabilities. Such techniques employ the entire set K of distributions consistent with the available information. They are thus compatible with either the strict Bayesian view that some member of K must be the "true" but unknown joint distribution or the alternative views of such authors as Levi (1974,1980), Kyburg (1987), and R.C. Jeffrey (1984,1987).

Let $P^n$ denote the (n-1)-dimensional simplex of n-component probability distributions over S. For problems of *partial uncertainty* of the type under discussion, what is known about a distribution p over S is that $p\in K$, where $K\subseteq P^n$. When $K=\{p\}$, the problem reduces to decision under risk, and when $K=P^n$, to decision under uncertainty. Thus, it would be reasonable to require that any technique for decision making under partial uncertainty reduce to well-justified techniques for the limiting cases (cf. Levi, 1980, p. 98).

Three techniques are discussed: Levi's, Gardenfors' (1979), and a generalization of Hurwicz' (1951a) pessimism–optimism criterion, which includes Gardenfors' as a special case.

### 3.1 Levi and Gardenfors

For Levi, an action $a_k$ is *E-admissible* iff there exists a $p\in K$ such that

$$\sum_{j=1}^{n} p(s_j)u_{kj} = \max_{i\in\{1,\dots,m\}} \sum_{j=1}^{n} p(s_j)u_{ij}.$$

If there is only one E-admissible action, it is selected. Otherwise (simplifying slightly), the information $p\in K$ is discarded and the maximin criterion is applied to the E-admissible actions.

This method has been criticized by Gardenfors and Sahlin (1982, 1987) as not sufficiently risk-aversive. To illustrate, let H denote the outcome "heads" when a particular coin is tossed, and let T denote "tails". Suppose K consists of all convex combinations of the distributions $p_1$ and $p_2$, where $p_1(H)=0.4$ and $p_2(H)=0.6$. For actions and utilities

284

|       | $s_1$ | $s_2$ |
|-------|-------|-------|
| $a_1$ | 1000  | -995  |
| $a_2$ | -995  | 1000  |
| $a_3$ | 0     | 0     |

a rational choice seems to be action $a_3$. However, under Levi's approach, although $a_1$ and $a_2$ are E-admissible, $a_3$ is not, and it would be rejected.

For the alternative proposed by Gardenfors, an action for which the minimum expected utility (as p ranges over K) is maximized is selected. Formally, select $a_k$ such that

$$\min_{p \in K} \sum_{j=1}^{n} p(s_j) u_{kj} = \max_{i \in \{1,\ldots,m\}} \min_{p \in K} \sum_{j=1}^{n} p(s_j) u_{ij}.$$

For the example above, $a_3$ is identified as the best action, since its minimum expected utility (over K) is 0, vs. -197 for both $a_1$ and $a_2$. Gardenfors' criterion specializes to maximin when $K = P^n$ (thus, it may be referred to as generalized maximin, abbreviated GM). When $K = \{p\}$, GM specializes to Bayesian expected utility maximization.

The set $U(a_k)$ of utilities associated with any $a_k$, defined as

$$U(a_k) = \{\sum_{j=1}^{n} p(s_j) u_{kj} | p \in K\},$$

is the range of a linear function of K: $f_k : P^n \to \mathbb{R}$, where $f_k(p) = \sum_{j=1}^{n} p(s_j) u_{kj}$. Thus, $U(a_k)$ is convex. Since $U(a_k) \subseteq \mathbb{R}$, $U(a_k)$ is an interval:

$$U(a_k) = [\min_{p \in K} \sum_{j=1}^{n} p(s_j) u_{kj}, \max_{p \in K} \sum_{j=1}^{n} p(s_j) u_{kj}].$$

Linear programming (minimizing and maximizing the value of $\sum_{j=1}^{n} p(s_j) u_{kj}$ subject to the constraints defining K) may be used to determine the endpoints of the interval associated with any $a_k$.

For the decision problem involving the randomly selected object, the utility intervals for each action are:

$$U(a_{BS}) = [-0.5, 4]$$
$$U(a_{BC}) = [0.1, 3.4]$$
$$U(a_{WS}) = [-50, 127]$$
$$U(a_{WC}) = [-1, 2.3].$$

Thus, action $a_{BC}$ is selected as optimal by GM.

Using the simplex algorithm, determination of the minimum expected utility for each action is reasonably efficient. The same number of applications of a linear programming algorithm suffice to determine the E-admissible set; however, for m actions, an additional m−1 equations and m−1 variables are required, assuming that the linear programming problem is expressed in "standard form" (Papadimitriou and Steiglitz, 1982).

Gardenfors and Sahlin (1982) criticize Levi's theory on grounds besides unrealistic optimism. They show that his technique violates the principle of *independence of irrelevant alternatives* (Luce and Raiffa, 1957); i.e., that addition of actions to a decision problem may make previously E-inadmissible actions E-admissible. It is also possible to alter the set of E-admissible actions by conjoining two or more states. (GM violates neither principle.)

While Levi's method may be insufficiently risk-aversive for certain situations, GM may be regarded as too pessimistic. Consider a decision problem involving two actions, $a_1$ and $a_2$, and a set of states S a probability distribution over which is known only to the extent that $p \in K$, giving rise to utility intervals

$$U(a_1) = [2.4, 56]$$
$$U(a_2) = [-3.2, 1.5]$$

Any reasonable criterion for decision making with utility intervals, including GM, would identify $a_1$ as the best act. Suppose instead that it is known that $p \in K'$, with utility intervals

$$U^*(a_1) = [4.8, 6.0]$$
$$U^*(a_2) = [4.7, 950]$$

GM would select $a_1$, whereas most people would probably be inclined to select $a_2$. For Levi, both acts are E-admissible. Maximin would be applied to attempt to decide between them. However, maximin is itself not immune to this type of difficulty. For the decision problem

|       | $s_1$ | $s_1$ |
|-------|-------|-------|
| $a_1$ | 4.8   | 6     |
| $a_2$ | 4.7   | 950   |

maximin picks $a_1$, although the maximum to be gained (over the payoff for choosing $a_2$) by this choice is 0.1 utiles vs. a maximum "loss" of $950 - 6 = 944$ utiles. It was to handle this type of situation that Savage developed his "minimax regret" criterion, but this criterion itself has a number of serious flaws (Luce and Raiffa, 1957).

### 3.2 Generalized Hurwicz Criterion

An alternative to both maximin and minimax regret for decision under uncertainty is Hurwicz' "pessimism-optimism" criterion (Hurwicz, 1951a; Luce and Raiffa, 1957): for some fixed $0 \leq \alpha \leq 1$, select an action $a_k$ such that

$$\max_{i \in \{1,\ldots,m\}} \left( \alpha \min_{j \in \{1,\ldots,n\}} u_{ij} + (1-\alpha) \max_{j \in \{1,\ldots,n\}} u_{ij} \right) = \alpha \min_{j \in \{1,\ldots,n\}} u_{kj} + (1-\alpha) \max_{j \in \{1,\ldots,n\}} u_{kj}.$$

When $\alpha = 1$, the pessimism-optimism index is the most pessimistic, and coincides with maximin. When $\alpha = 0$, the index is the most optimistic, selecting an action with the best "best outcome". For values $0 < \alpha < 1$, the index is more or less conservative, depending on how close $\alpha$ is to 1.

An extension to problems of decision making under partial uncertainty (suggested in Hurwicz, 1951b) is the generalized Hurwicz criterion (GH): for some $0 \leq \alpha \leq 1$, select $a_k \in A$ such that

$$\alpha(\min U(a_k)) + (1-\alpha)(\max U(a_k))$$

is maximized. The parameter $\alpha$ selects a point along the utility interval by which to compare actions. When $\alpha = 1$, GH is most pessimistic, and coincides with GM. The ability to adjust the degree of risk-aversiveness in this way seems to be an advantage. For the problem above involving utility intervals

$$U^*(a_1) = [4.8, 6.0]$$
$$U^*(a_2) = [4.7, 950],$$

GH prescribes the choice of $a_2$ for any value of $\alpha$ less than $944/944.1$. (It can also be shown that GH behaves like GM with respect to augmentation of the set of actions under consideration and repartitioning of states.) However, use of GH is computationally more difficult than use of GM, for $0 < \alpha < 1$.

## 4. INTERVAL-VALUED PROBABILITIES

Partial uncertainty has also been represented by interval-valued probabilities of states (Wolfenson and Fine, 1982; Loui et al., 1986; Kyburg, 1987). These may be confidence intervals resulting from sampling, imprecisely stated subjective probabilities, etc. Any specification of interval probabilities over S has an associated largest convex set, K, of (real-valued) distributions consistent with it. Let $l_j$ and $u_j$ denote the endpoints of the probability interval for state $s_j$. Then K is the intersection of $P^n$ with the set of solutions to the system of inequalities

$$p(s_1) \leq u_1$$
$$p(s_1) \geq l_1$$
$$\cdots$$
$$p(s_n) \leq u_n$$
$$p(s_n) \geq l_n.$$

Since K is the intersection of two convex sets, K is itself convex, and gives rise to expected utility intervals for each act. These are determined by adding the equation $\sum_{j=1}^{n} p(s_j) = 1$ to the inequalities and maximizing and minimizing (via linear programming) the function $\sum_{j=1}^{n} p(s_j) u_{kj}$ for each act $a_k$.

286

Thus, any of the decision techniques proposed for use with convex sets of (real-valued) probability distributions over X may be used also with probability intervals.

## 5. STRATEGIES FOR IMPROVING DECISIONS

For the method of Loui et al. (1986), an ordering among the actions in A is defined as a>a' iff min $U(a) > \max U(a')$. If there is a unique maximal element under this ordering in the set A, then that action is selected (as is the case with Levi's method, GH, for any value of $\alpha$, and thus GM). Otherwise, it is attempted to reduce the width of the utility intervals $U(a_i)$ by reducing the size of the probability intervals $[l_j, u_j]$. It is proposed that this be achieved by adopting "a more relaxed attitude toward error"; for example, through the use of .90 vs. .95 confidence levels in constructing the intervals.

When information on joint probabilities is available in the form of known marginal probabilities, the spaces on which the marginals are defined and the manner in which they are processed can affect the width of the resulting $U(a_i)$ and thus the quality of the decision, regardless of the criterion used. (The results discussed below can be extended to apply to real- or interval-valued distributions in any combination, allowing a hierarchical type of analysis.)

Suppose V is a set of variables (e.g., {color, shape}) each of which has a finite set of values (e.g., {black, white}, {spherical, cylindrical}) the Cartesian product of which ({(black, spherical),...,(white, cylindrical)}) may be interpreted as the set S of states in a decision problem. A *model* of a set of variables $V=\{v_1,\ldots,v_w\}$ is a set $X=\{V_1,\ldots,V_m\}$ such that $\bigcup_{j=1}^{m} V_j \subseteq V$ and $V_i \not\subset V_j$ for all $i,j \in \{1,...,m\}$. (X needn't be a cover of V.) The algebra of models and its relevance to data analysis have been studied by a number of authors (Cavallo and Klir, 1979; Lee, 1983; Edwards and Havranek, 1985).

### 5.1 Choose Coarser Models

A fundamental result is that the coarser the structure of a model over which marginals are defined, the smaller the set of joint distributions compatible with them.

*Definition*: Let p be a distribution defined on a set of states S, where S is the Cartesian product of the sets of values of variables in $V=\{v_1,\ldots,v_w\}$. *Projecting* p onto $Z \subseteq V$ results in its associated (unique) marginal distribution over Z, denoted $\pi_Z(p)$. For example, projecting

| C | S | p*(.) |
|---|---|---|
| B | S | 0.42 |
| B | C | 0.28 |
| W | S | 0.18 |
| W | C | 0.12 |

onto {C} gives the marginal distribution

| C | $\pi_{\{C\}}(p^*)(.)$ |
|---|---|
| B | 0.7 |
| W | 0.3 |

Projecting p onto a model $X=\{V_1,\ldots,V_m\}$ of V yields the set of marginals $\pi_X(p)=\{\pi_{V_i}|i=1,...,m\}$.

*Definition*: *Extension* of a set of marginals $P=\{p_1,\ldots,p_m\}$ over a model $X=\{V_1,\ldots,V_m\}$ of V results in the convex polyhedron $E_V(P)$ of joint distributions (extensions) over V (with associated $|S|=n$) compatible with them:

$$E_V(P) = \{p \in P^n | \pi_{V_i}(p) = p_i, i=1,...,m\}.$$

If $V = V_1, \ldots, V_m$, then $E_V(P)$ may be written $E(P)$. For any $p \in P^n$, $p \in E_V(\pi_X(p))$.

*Definition* (Cavallo and Klir, 1979): For models X and Y of set V, X is a *refinement* of Y, denoted $X \leq Y$, iff for each $V_x \in X$ there exists a $V_y \in Y$ such that $V_x \subseteq V_y$. For example, $\{\{C\},\{M,S\}\}$ is a refinement of $\{\{C,M\},\{M,S\},\{C,S\}\}$.

*Fact* (Higashi, 1984): $X \leq Y$ implies $E_V(\pi_Y(p)) \subseteq E_V(\pi_X(p))$.

Thus, if X<Y, then it is preferable (all else being equal) to obtain marginals over Y, since $E_V(\pi_Y(p)) \subseteq E_V(\pi_X(p))$ implies (among other things) that the probability and utility intervals associated with $E_V(\pi_Y(p))$ are subintervals of those associated with $E_V(\pi_X(p))$. (Unfortunately, not all models of a given set of variables are comparable under $\leq$; e.g., neither $\{\{C,M\},\{M,S\}\}$ nor

287

{{C,S},{S,M}} is a refinement of the other. It cannot be determined in advance of obtaining data over incomparable models which of them gives more information about a particular joint distribution.)

The most refined model, $\{\emptyset\}$, provides no information: $E_V(\pi_{\{\emptyset\}}(p))=P^n$. The least refined model over V, {V}, gives a unique joint distribution. However, the less refined the model, the more difficult it may be to obtain probabilities. In such cases there is a trade-off between reduction of the size of $E_V(P)$ and the effort required to obtain probabilities over large groups of variables simultaneously.

### 5.2 Refine and Extend, then Marginalize

(This subsection elaborates a concept discussed in Cavallo and Pittarelli, 1987b.)

It is not always the case that every variable contained in the elements of a model $X = \{V_1,...,V_m\}$ is perceived as relevant to a particular decision problem. Yet, marginals over elements of such a model may be the only source of information regarding a distribution over the set of variables of actual interest. This may happen when partial studies of a phenomenon that is no longer observable were carried out over variables some of which are no longer considered important, when a pre-existing monitoring scheme that it would be too expensive or time-consuming to alter includes more variables than it is currently desired to observe, etc. The second basic result to be presented in this section is applicable to this type of problem.

Suppose that marginal distributions over a model $X=\{V_1,\ldots,V_m\}$ are known. They are assumed to be marginals of some $p_z$; i.e., the marginals are consistent. Suppose that the variables of interest (e.g., for decision making) are the elements of $V_o \subseteq V = V_1 \cup \cdots \cup V_m$. The probability distribution of interest is then $p_o = \pi_{V_o}(p_z)$. If information is given as $p_z \in A$, then $p_o \in \pi_{V_o}(A)$, the range of the function $\pi_{V_o}$ applied to the set A.

- If $V_o \in X$, then $p_o$ (the unique distribution over $V_o$ consistent with the information $\pi_{V_1}(p_z),...,\pi_{V_m}(p_z))$ is immediately given as $\pi_{V_o}(p_z)$.
- If $V_o \subset V_k \in X$, then $p_o = \pi_{V_o}(\pi_{V_k}(p_z))$, since $V_i \subseteq V_j$ implies $\pi_{V_i}(\pi_{V_j}(p)) = \pi_{V_i}(p)$.
- Otherwise, there are two extremes:

(1) $p_o \in \pi_{V_o}(E_V(\pi_X(p_z)))$.
(2) Let $X_o = \{\{v\} | v \in V_o\}$. Then $p_o \in E_{V_o}(\pi_{X_o}(p_z)) = \pi_{V_o}(E_V(\pi_{X_o}(p_z)))$.

Since $X_o \leq X$, $E_V(\pi_X(p_z)) \subseteq E_V(\pi_{X_o}(p_z))$, which implies that
$\pi_{V_o}(E_V(\pi_X(p_z))) \subseteq \pi_{V_o}(E_V(\pi_{X_o}(p_z)))$; i.e., that $\pi_{V_o}(E(\pi_X(p_z))) \subseteq E(\pi_{X_o}(p_z))$. $X_o$ and X are, respectively, the most and least refined models of the set of variables for which information (in the form of marginal distributions) is available, and which cover $V_o$. For any covering model W, $X_o \leq W$. If also $W \leq X$, then it is guaranteed that $(p_o=) \pi_{V_o}(p_z) \in \pi_{V_o}(E_V(\pi_W(p_z)))$. $\pi_{V_o}(E_V(\pi_X(p_z)))$ is the smallest of these sets guaranteed to contain $p_o$, and its associated probability and utility intervals are therefore the narrowest. However, computation based directly on (1), above, could be unnecessarily expensive. When $V_o \neq V$, the same set (intervals) can sometimes be obtained less expensively.

*Definition* (Fagin, 1983): A *path* from element $V_1$ to $V_k$ in X is a sequence of elements $V_1,\ldots,V_k$ such that $V_i \cap V_{i+1} \neq \emptyset$, for $1 \leq i < k$. Two elements of X are *connected* iff there is a path from one to the other. A set of elements is connected iff each pair is connected. A *connected component* of X is a connected set of elements of X such that no proper subset is connected.

Any model X can be represented as a unique partition, $X_p$, into connected components. E.g.,
$$X = \{\{A,B\},\{B,C\},\{D,E\},\{E,F\}\}$$
can be represented as
$$X_p = \{\{\{A,B\},\{B,C\}\},\{\{D,E\},\{E,F\}\}\}.$$

*Definition*: For a given X and $V_o$, let $X/V_o$ denote the subset of members of X contained in connected components involving elements of $V_o$:
$$X/V_o = \{\cup_{C \in X_p} | \text{for some } V_i \in C, V_i \cap V_o \neq \emptyset\}.$$

*Definition*: A *channel* in a model X between variables $v_1$ and $v_k$ is a sequence of distinct variables
$$(v_1, v_2, \ldots, v_k), k \geq 3,$$
such that, for $j=2,...,k-1$, there exist $V_a, V_b \in X$ such that $V_a \neq V_b$, $\{v_j, v_{j-1}\} \subseteq V_a$ and $\{v_j, v_{j+1}\} \subseteq V_b$.

288

*Claim*: W, the model resulting from application of the algorithm below (adapted from Maier and Ullman, 1982) to X and $V_o$, is the most refined model for which $\pi_{V_o}(E(\pi_X(p_z))) = \pi_{V_o}(E(\pi_W(p_z)))$:

1. $W \leftarrow X/V_o$.

Apply the following two steps in any order until neither has any effect on the current value of W:

2. If a variable $v \notin V_o$ appears in only one element of W, remove v from that element.
3. If W contains an element $V_i$ and an element $V_j$ such that $V_i \subset V_j$, then $W \leftarrow W - \{V_i\}$. □

The algorithm removes from X all $v \notin V_o$ that are not elements of some channel between some pair of variables in $V_o$. (When X covers $V_o$, $X_o \leq W \leq X$. It is applicable also when X does not cover $V_o$, in which case $\{\emptyset\} \leq W \leq X$; e.g., $V_o = \{A,B,C\}$, $X = \{\{A,D\},\{D,B,M\},\{E,F,G,H,M\}\}$, $W = \{\{A,D\},\{D,B\}\}$.)

Recall the decision problem involving the shape and color of a randomly selected object. Suppose that information is also available regarding what it is made of (aluminum or plywood), and its density (low, medium, high), unit price (low, high), and attractiveness (low, medium, high), in the form of three probability distributions

| C | M | $p_1(.)$ | M | S | D | $p_2(.)$ | U | A | $p_3(.)$ |
|---|---|---|---|---|---|---|---|---|---|
| B | A | 0.6 | A | S | L | 0.4 | L | L | 0.1 |
| B | P | 0.1 | A | S | M | 0.0 | L | M | 0.0 |
| W | A | 0.2 | A | S | H | 0.0 | L | H | 0.0 |
| W | P | 0.1 | A | C | L | 0.4 | H | L | 0.2 |
| | | | A | C | M | 0.0 | H | M | 0.3 |
| | | | A | C | H | 0.0 | H | H | 0.4 |
| | | | P | S | L | 0.0 | | | |
| | | | P | S | M | 0.2 | | | |
| | | | P | S | H | 0.0 | | | |
| | | | P | C | L | 0.0 | | | |
| | | | P | C | M | 0.0 | | | |
| | | | P | C | H | 0.0 | | | |

For this problem, $X = \{\{C,M\},\{M,S,D\},\{U,A\}\}$ and $V_o = \{C,S\}$. The simplest approach is to follow (2), above, extracting the distributions (marginals of marginals of $p_z$) $\pi_{\{C\}}(p_1)$ and $\pi_{\{S\}}(p_2)$ and forming their extension over $V_o$, $E(\pi_{X_o}(p_z))$. It is guaranteed that $p_o \in E(\pi_{X_o}(p_z))$; since $E(\pi_{X_o}(p_z))$ is a proper subset of $P^4$, some information is gained. But the sharpest constraints on $p_o$ derivable from the data, following (1), above, are given by (the linear equations corresponding to) forming $E(\pi_X(p_z))$ (a subset of $P^{144}$) and projecting onto $P^4$. The algorithm above identifies $W = \{\{C,M\},\{M,S\}\}$. Thus, the same (sharpest derivable) constraints on $p_o$ are given by (the linear equations for) constructing $E(\pi_W(p_z))$ (a subset of $P^8$) and projecting onto $P^4$.

The utility intervals determined by $E(\pi_{X_o}(p_z))$ are (as calculated previously):

$U(a_{BS}) = [-0.5, 4]$

$U(a_{BC}) = [0.1, 3.4]$

$U(a_{WS}) = [-50, 127]$

$U(a_{WC}) = [-1, 2.3]$.

Those determined by $\pi_{V_o}(E(\pi_W(p_z)))$ $(= \pi_{V_o}(E(\pi_X(p_z))))$ are sharper:

$U'(a_{BS}) = [-0.5, 2.5]$

$U'(a_{BC}) = [1.2, 3.4]$

$U'(a_{WS}) = [9, 127]$

$U'(a_{WC}) = [-1, 1.2]$.

With intervals U', since the minimum expected utility associated with $a_{WS}$ exceeds the maximum for any other action, $a_{WS}$ would be selected using any of the techniques discussed. Thus, the (hypothetical) object's material conveys useful information regarding the joint distribution of color and shape. On the other hand, while the other attributes could not be said to be irrelevant, there would be no way to determine their effects on color and shape from the information given.